\documentclass[10pt,twocolumn,letterpaper]{article}

\usepackage[pagenumbers]{iccv}      % To produce the REVIEW version
\definecolor{iccvblue}{rgb}{0.21,0.49,0.74}
\usepackage[pagebackref,breaklinks,colorlinks,allcolors=iccvblue]{hyperref}
\usepackage{xcolor}
\usepackage{booktabs}
\usepackage{subcaption}
\usepackage{graphicx}
\usepackage{caption}
\usepackage{amssymb}

\usepackage[linesnumbered,ruled,vlined]{algorithm2e}

\def\paperID{14} % *** Enter the Paper ID here
\def\confName{ICCV CDEL}
\def\confYear{2025}

\title{Class-Proportional Coreset Selection for Difficulty-Separable Data}

\author{Elisa Tsai\\
University of Michigan\\
{\tt\small eltsai@umich.edu}
\and
Haizhong Zheng\\
Carnegie Mellon University\\
{\tt\small haizhonz@cmu.edu}
\and
Atul Prakash\\
University of Michigan\\
{\tt\small aprakash@umich.edu}
}

\begin{document}
\maketitle
\begin{abstract}
High-quality training data is essential for building reliable and efficient machine learning systems. One-shot coreset selection addresses this by pruning the dataset while maintaining or even improving model performance, often relying on training-dynamics-based data difficulty scores. However, most existing methods implicitly assume class-wise homogeneity in data difficulty, overlooking variation in data difficulty across different classes.
In this work, we challenge this assumption by showing that, in domains such as network intrusion detection and medical imaging, data difficulty often clusters by class. We formalize this as class-difficulty separability and introduce the Class Difficulty Separability Coefficient (CDSC) as a quantitative measure.  We demonstrate that high CDSC values correlate with performance degradation in class-agnostic coreset methods, which tend to overrepresent easy majority classes while neglecting rare but informative ones.
To address this, we introduce class-proportional variants of multiple sampling strategies. Evaluated on five diverse datasets spanning security and medical domains, our methods consistently achieve state-of-the-art performance. For instance, on CTU-13, at an extreme $99\%$ pruning rate,  a class-proportional variant of Coverage-centric Coreset Selection (CCS-CP) shows remarkable stability, with accuracy dropping only $2.58\%$, precision $0.49\%$, and recall $0.19\%$. In contrast, the class-agnostic CCS baseline, the next best method, suffers sharper declines of $7.59\%$ in accuracy, $4.57\%$ in precision, and $4.11\%$ in recall.
We further show that aggressive pruning enhances generalization in noisy, imbalanced, and large-scale datasets. Our results underscore that explicitly modeling class-difficulty separability leads to more effective, robust, and generalizable data pruning, particularly in high-stakes scenarios such as security and medical imaging.

\end{abstract}

% Old abstract with NIDS focus
%Network intrusion detection systems (NIDS) are essential for defending modern infrastructure against evolving cyber threats. However, training effective machine learning models for NIDS typically requires large-scale datasets to capture the complex and diverse nature of real-world network traffic. In this work, we explore whether learning-dynamics-based coreset selection can reduce training set size without compromising detection performance.
%We systematically evaluate multiple coreset strategies---including hardest-example sampling, Coverage-centric Coreset Selection (CCS), and our proposed class-balanced variants---across three widely used NIDS benchmarks: CICIDS2017, UNSW-NB15, and CTU-13. Our experiments reveal that a class-balanced CCS method achieves remarkable data efficiency: in CICIDS2017, training with just 1\% of the original data incurs less than 0.5\% performance degradation, and even with 0.1\% original training data, the test accuracy decreases by just $\sim$ 1\%. 
%Moreover, echoing findings from the previous literature, we observe that removing up to 70\% of the training data can improve generalization, indicating the presence of detrimental examples. These results highlight coreset selection as a powerful tool not only for data compression but also for dataset denoising in security applications.     
\section{Introduction}
\label{sec:intro}
\begin{figure*}[htbp]
  \centering

  % Second row: CIFAR10
  \begin{subfigure}{0.21\textwidth}
    \includegraphics[width=\linewidth]{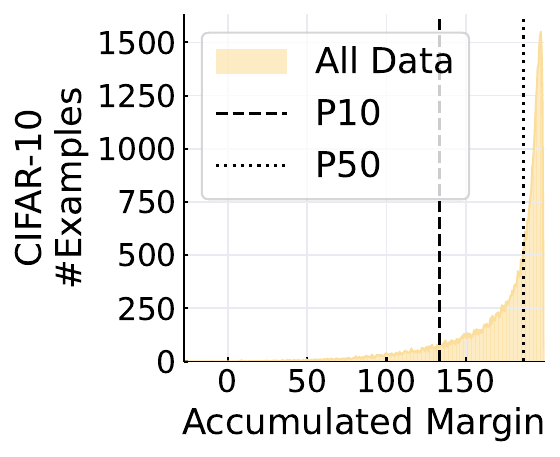}
  \end{subfigure}
  \begin{subfigure}{0.18\textwidth}
    \includegraphics[width=\linewidth]{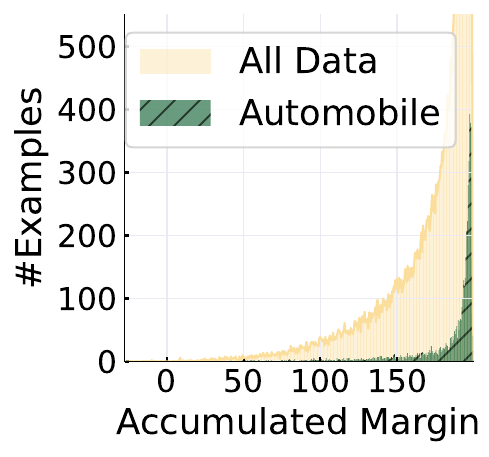}
  \end{subfigure}
  \begin{subfigure}{0.18\textwidth}
    \includegraphics[width=\linewidth]{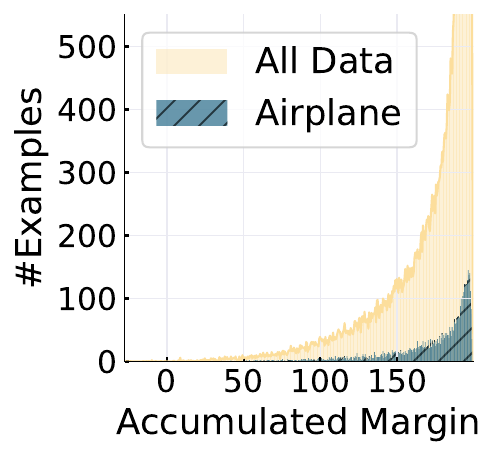}
  \end{subfigure}
  \begin{subfigure}{0.18\textwidth}
    \includegraphics[width=\linewidth]{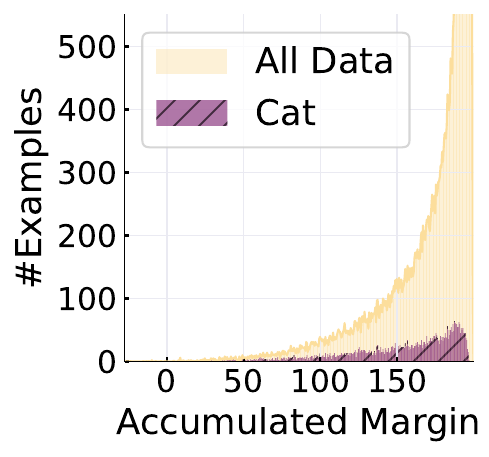}
  \end{subfigure}
  \begin{subfigure}{0.18\textwidth}
    \includegraphics[width=\linewidth]{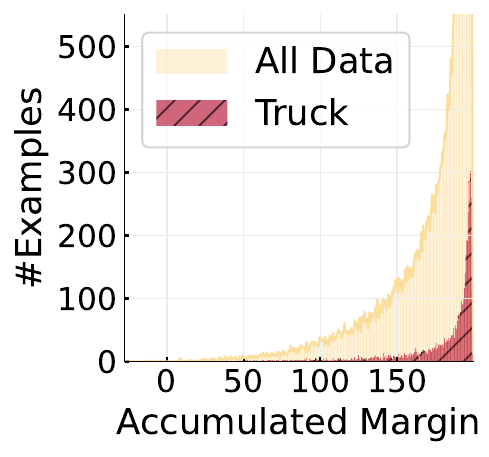}
  \end{subfigure}
  % First row: CICIDS2017
  
  \hspace{4mm}\begin{subfigure}{0.19\textwidth}
    \includegraphics[width=\linewidth]{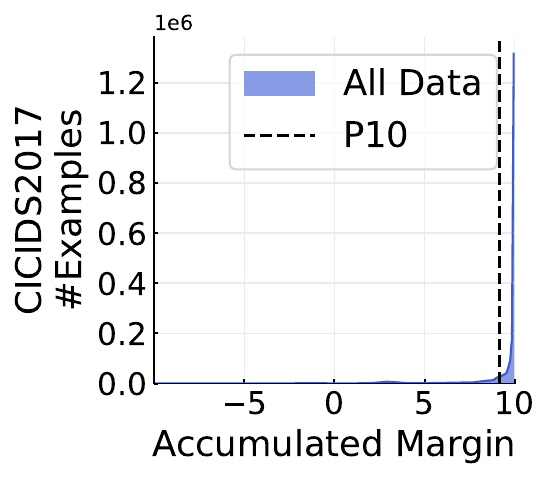}
  \end{subfigure}
  \begin{subfigure}{0.18\textwidth}
    \includegraphics[width=\linewidth]{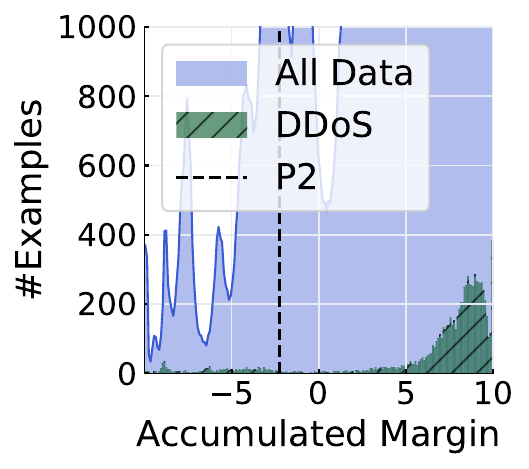}
  \end{subfigure}
  \begin{subfigure}{0.18\textwidth}
    \includegraphics[width=\linewidth]{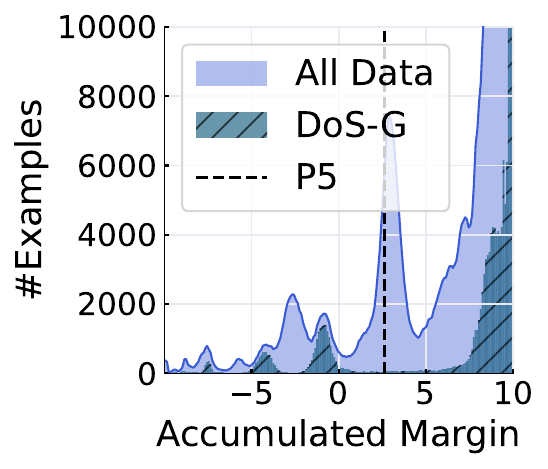}
  \end{subfigure}
  \begin{subfigure}{0.18\textwidth}
    \includegraphics[width=\linewidth]{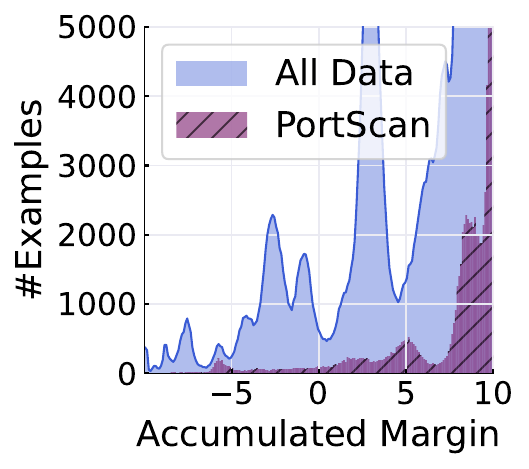}
  \end{subfigure}
  \begin{subfigure}{0.18\textwidth}
    \includegraphics[width=\linewidth]{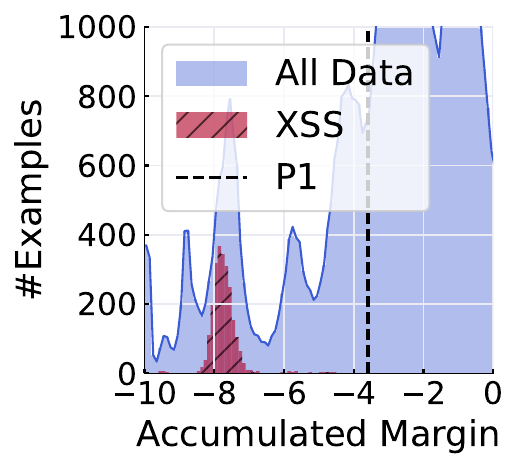}
  \end{subfigure}

  \caption{
    Distribution of accumulated margin (AUM) score for CIFAR-10 and CICIDS2017. Higher AUM values indicate ``easier'' examples, as inferred from training dynamics.  \textbf{Top Row:} CIFAR-10 dataset (200 epochs). \textbf{Bottom Row:} CICIDS2017 dataset (20 epochs). Percentile thresholds (1\%, 2\%, 5\%, 10\%, and 50\%) are indicated as vertical dashed lines (P1, P2, P5, P10, and P50).
  }

  \label{fig:aum_combined}
\end{figure*}

Coreset selection and data pruning have emerged as effective tools for reducing the size of training datasets while preserving, and in some cases even improving, model performance. This paradigm has seen success across vision tasks such as image classification and object detection~\cite{toneva2018empirical, paul2021deep, coleman2019selection, zheng2022coverage, zheng2024bridging, maharana2023d2, xia2022moderate, choi2023bws, zheng2024elfs, lee2024coreset, killamsetty2021retrieve}, as well as recent applications in large language model (LLM) pretraining and supervised fine-tuning~\cite{xia2024rethinking, xia2024less, choe2024your, liu2024less, zheng2024learn, joaquin2024in2core}. However, despite this growing body of work, the underlying assumptions behind many coreset selection methods remain largely unexamined.

State-of-the-art coreset selection methods for classification tasks~\cite{zheng2022coverage, maharana2023d2, choi2023bws, zheng2024elfs} typically perform sampling based on data difficulty scores derived from training dynamics such as loss, forgetting events, or confidence margin~\cite{pleiss2020identifying, toneva2018empirical, paul2021deep}. These methods are \textit{class-agnostic} and rely on a key simplifying assumption that \textit{each class exhibits a similar data difficulty distribution}. In other words, they assume difficulty is comparably distributed across classes, enabling ranking and selection of examples without considering class membership. This assumption simplifies sampling and supports general-purpose pruning strategies that perform well on balanced datasets like CIFAR-10/100~\cite{krizhevsky2009learning} and ImageNet~\cite{deng2009imagenet}, where inter-class data difficulty divergence is minimal.

However, in many real-world domains—such as cybersecurity~\cite{cicids2017, ctu13, unswnb15} and medical diagnostics~\cite{medmnist}—the assumption of homogeneous difficulty distribution across classes breaks down. Some classes are consistently easier to learn than others due to differences in prevalence, label quality, or intra-class variability.
In network intrusion detection, for example, benign traffic is abundant, highly redundant, and generally easy to classify. In contrast, attack traffic is relatively rare, harder to learn, and often exhibits high intra-class variability. 

As shown in Figure~\ref{fig:aum_combined}, this leads to tightly overlapping AUM~\cite{pleiss2020identifying} distributions across CIFAR-10 classes, but clearly separable clusters in CICIDS2017, a large-scale network intrusion detection dataset, where certain attack types (e.g., DoS-GoldenEye, XSS) form distinct modes in difficulty space.
A similar effect is observed in medical imaging. Common conditions are typically learned early during training, while rare diseases or ambiguous cell types persist as hard examples throughout. In both domains, difficulty is not evenly distributed but tends to cluster by classes. We refer to this phenomenon as \textit{class-difficulty separability}.

To characterize the extent of difficulty variation across classes, we introduce the \textit{Class Difficulty Separability Coefficient} (CDSC, Section~\ref{ssec:separability}), a normalized Jensen–Shannon divergence~\cite{lin2002divergence} computed over class-conditional data difficulty distributions. CDSC quantifies how distinctly different the difficulty scores are across classes. In Section~\ref{sssec:cdsc}, we show that class-agnostic coreset selection methods degrade more significantly on datasets with higher $\delta_{\text{CD}}$. This supports our hypothesis that modeling class-specific difficulty becomes increasingly important as difficulty separability increases.

In this work, we investigate how class-difficulty separability impacts the effectiveness of one-shot coreset selection and propose a simple yet effective remedy: \textit{class-proportional} variants of existing sampling strategies. Specifically, we adapt several data difficulty–based methods, including hardest example~\cite{pleiss2020identifying, hong2024evolution}, sliding-window~\cite{choi2023bws, zheng2024elfs}, and Coverage-centric Coreset Selection (CCS)~\cite{zheng2022coverage}, by applying coverage within each class and allocating selection quotas proportionally to the number of examples in each class. We show that this modification consistently improves the performance of each method. Our approach is modular and can be seamlessly integrated with other training-dynamics-based selectors.

We evaluate our method on five datasets across security and medical domains: CICIDS2017~\cite{cicids2017}, CTU-13~\cite{ctu13}, UNSW-NB15~\cite{unswnb15}, BloodMNIST, and DermaMNIST~\cite{medmnist}. These datasets exhibit much greater class-wise difficulty heterogeneity than standard benchmarks like CIFAR-10/100~\cite{krizhevsky2009learning}. Our class-proportional methods consistently outperform class-agnostic baselines, especially at high pruning rates. CCS-CP achieves the best overall test accuracy, precision, and recall across all datasets. On CTU-13, CCS-CP retains performance even after pruning 90\% of training data, and at a 99\% pruning rate, it outperforms class-agnostic CCS by 5\% in accuracy, 4.6\% in precision, and 4.1\% in recall. Similar patterns hold throughout, with class-aware sampling offering stronger robustness under aggressive pruning.  Moreover, we observe that on large and noisy datasets, aggressive pruning can improve generalization. This aligns with prior findings that some examples---either too easy and redundant, or too hard and noisy---may hinder learning. Removing such examples allows the model to focus on informative samples, resulting in better test performance. These results highlight the importance of modeling class-difficulty separability when designing data-efficient learning pipelines. 

In summary, our key contributions are:

\begin{itemize}
    \item We identify and formalize \textit{class-difficulty separability} as a structural property that can degrade the performance of class-agnostic coreset selection methods. To quantify it, we introduce the \textit{Class Difficulty Separability Coefficient (CDSC)}, a normalized Jensen–Shannon divergence over class-wise difficulty distributions. Higher CDSC values indicate greater performance gaps between class-agnostic and class-aware methods.
    
    \item We propose \textit{class-proportional variants} of sampling strategies that allocate budgets across classes and sample based on within-class difficulty to enhance representativeness and robustness.
    
    \item We evaluate our methods on five datasets from security and medical domains, demonstrating that class-proportional sampling achieves state-of-the-art data efficiency and maintains or improves generalization, even after pruning up to 90\% of training data.
\end{itemize}

\section{Related Work}

\subsection{Coreset Selection}

Given a fixed budget, coreset selection seeks an optimal subset of training data that maximizes model performance, which can match or even surpass the performance achieved using the full dataset. Existing methods fall into two main categories: (1) training-free methods, such as geometry or diversity-based selection~\citep{chen2012super, welling2009herding, prototypicality, xia2022moderate}; and (2) training-dynamics–based methods~\cite{pleiss2020identifying, toneva2018empirical, paul2021deep}, which use training behavior to estimate data importance. We focus on the latter due to its better empirical performance.

\noindent\textbf{Data Difficulty Metrics.}
Training dynamics–based coreset selection methods estimate data importance by analyzing how examples are learned over time. Two widely used difficulty metrics are:

\begin{itemize}
    \item \textbf{AUM}~\citep{pleiss2020identifying} (Area Under the Margin): Measures the accumulated logit margin between the correct and top competing class over epochs, reflecting prediction confidence.
    \item \textbf{Forgetting}~\citep{toneva2018empirical}: Counts how often a sample, once correctly classified, flips from correct to incorrect during training, highlighting unstable but informative examples.

\end{itemize}

\noindent Other training dynamics–based metrics include EL2N~\citep{paul2021deep} for early error magnitude, entropy~\citep{coleman2019selection} for prediction uncertainty, TDDS~\citep{zhang2024spanning} for depth-time dynamics, and EVA~\cite{hong2024evolution} for error fluctuation across training stages.

\noindent\textbf{Data Diversity Sampling}. 
Early coreset methods~\citep{toneva2018empirical, paul2021deep, coleman2019selection} prioritized hard examples, motivated by the SVM intuition that such points near decision boundaries are more informative~\cite{sener2017active}. However, selecting only hard examples degrades performance at high pruning rates~\citep{zheng2022coverage}, as these often lie in low-density regions. In contrast, easy examples represent high-density areas. Recent work~\citep{zheng2022coverage, maharana2023d2, xia2022moderate, choi2023bws, zheng2024elfs} improves upon this by sampling more strategically from the data difficulty distribution, removing both overly easy and overly hard examples, and emphasizing diversity. These methods have demonstrated that balancing difficulty and diversity leads to better performance across a wide range of pruning rates.

\subsection{Difficulty-Separable Data}

Machine learning has become a key component in addressing a wide range of security challenges~\cite{tsai2025harmful, prakash2024detecting, tsai2024terms, tsai2024modeling}.
Many real-world datasets exhibit inter-class difficulty divergence—some classes are consistently easier to learn, while others are inherently harder due to factors such as greater intra-class variability, ambiguous boundaries, or lower representation. This challenges standard, class-agnostic coreset selection methods. We highlight two domains where this structure is especially prominent:

\noindent\textbf{Network Intrusion Detection} Network intrusion datasets are typically dominated by benign traffic~\cite{ctu13, cicids2017, unswnb15}, with attack types that vary greatly in frequency and behavior. While coreset selection has been well studied in vision tasks, its application to intrusion detection remains limited. Some early work explores Bayesian coresets~\cite{zennaro2019analyzing} and feature selection for efficiency, but systematic evaluations of data pruning in this domain are scarce.

\noindent\textbf{Medical Imaging}. Medical image datasets~\cite{medmnist} often suffer from class imbalance and heterogeneous visual patterns within rare disease categories. EVA~\cite{hong2024evolution} is a training-dynamics–based metric to identify informative examples, achieving high compression rates with minimal accuracy loss. Few studies have systematically explored coreset selection in domains with strong inter-class difficulty divergence.

%\textbf{Network Intrusion Detection}. Network intrusion detection (NID) is vital for identifying malicious activities within network traffic. Traditional signature-based approaches~\cite{roesch1999snort, kumar2020integrated}, relying on predefined rules, often fail to generalize novel or subtle intrusions. Machine learning (ML)-based NID systems address these limitations by automatically learning features directly from network data~\cite{sommer2010outside, sharafaldin2018toward, farid2010combining, garcia2014empirical, shone2018deep, disha2022performance, wu2020pelican, ullah2024ids}. Recent advances include decision trees~\cite{farid2010combining, disha2022performance}, deep learning models such as ResNet~\cite{wu2020pelican}, LSTMs~\cite{wei2023xnids, gwon2019network}, autoencoders~\cite{mirsky2018kitsune, wei2023xnids}, and transformers~\cite{wu2022rtids, ullah2024ids}, which effectively capture complex and temporal traffic dynamics. Widely adopted benchmark datasets like CICIDS2017~\cite{cicids2017}, UNSW-NB15~\cite{unswnb15}, and CTU-13~\cite{ctu13} offer labeled network traffic with diverse protocols and modern attack scenarios, enabling rigorous evaluation of NID techniques.

\section{Methodology}

\begin{figure}[t]
\centering

% Row of three plots in one column
\begin{subfigure}[b]{0.3\columnwidth}
  \centering
  \includegraphics[width=\linewidth]{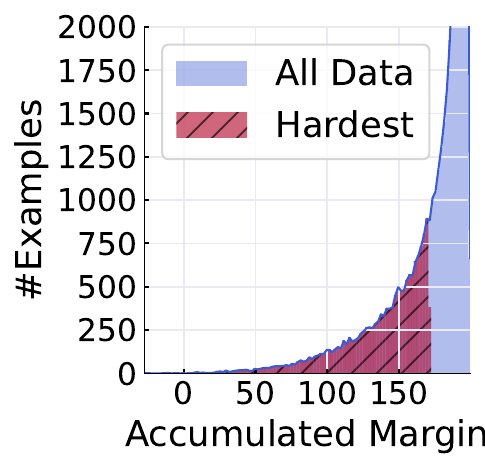}
  \caption{Hardest}
  \label{fig:cifar10-hardest}
\end{subfigure}
\hfill
\begin{subfigure}[b]{0.3\columnwidth}
  \centering
  \includegraphics[width=\linewidth]{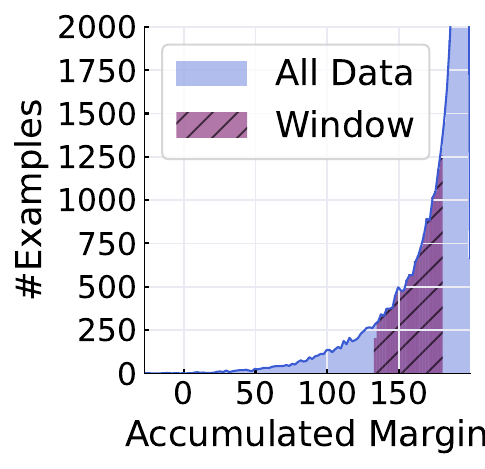}
  \caption{Sliding Window}
  \label{fig:cifar10-window}
\end{subfigure}
\hfill
\begin{subfigure}[b]{0.3\columnwidth}
  \centering
  \includegraphics[width=\linewidth]{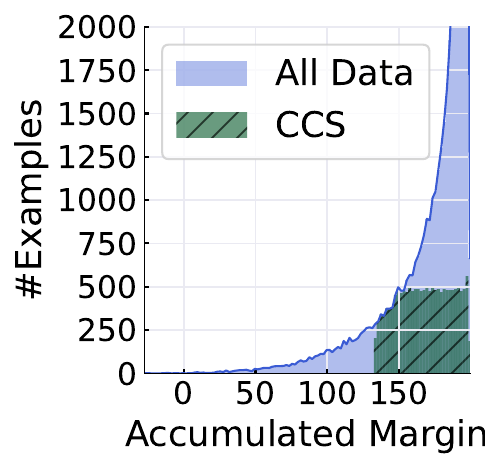}
  \caption{Stratified}
  \label{fig:cifar10-ccs}
\end{subfigure}

\vspace{-0.5em}
\caption{Demonstration of coresets selected on CIFAR-10 using different sampling strategies on AUM distribution with a 30\% coreset ratio.}
\label{fig:cifar10-comparison}
\end{figure}

In this section, we describe one-shot coreset selection and motivate the use of class-proportional sampling to address highly skewed class-wise data difficulty distributions.

\subsection{Problem Formulation}

Given a dataset $\mathcal{D} = \{x_1, x_2, \ldots, x_N\}$ drawn from an unknown distribution $P$, and a model $h_\mathcal{S}$ trained on a subset $\mathcal{S} \subset \mathcal{D}$, our goal is to select a coreset $\mathcal{S}$ of size $k$ that minimizes the expected loss on unseen data. Formally, the one-shot coreset selection problem is defined as:

\begin{equation}
  \mathcal{S}^* = \mathop{\mathrm{arg\,min}}\limits_{\mathcal{S} \subset \mathcal{D},\, |\mathcal{S}| = k } \mathbb{E}_{(x, y) \sim P}\left[\ell(x, y; h_\mathcal{S})\right],
\end{equation}

where $\ell(x, y; h_\mathcal{S})$ denotes the loss of model $h_\mathcal{S}$ on example $(x, y)$, and $k \leq N$ is the selection budget.

\noindent \textbf{Goals}. Coreset selection aims to identify a subset of the training data that maximizes model performance under a fixed size budget. We also aim to improve generalization by filtering out low-quality samples---ones that are mislabeled, noisy, or ``forgettable''---that may lead to overfitting or degrade performance.

\subsection{Motivation}

State-of-the-art coreset selection methods often operate by sampling from the data difficulty distribution, which is typically measured using training dynamics such as AUM~\citep{pleiss2020identifying} or forgetting score~\citep{toneva2018empirical}, to construct compact and diverse subsets. While early approaches prioritized the hardest examples~\citep{toneva2018empirical, paul2021deep, sener2017active}, this strategy can lead to performance degradation at high pruning rates~\citep{zheng2022coverage}. As shown in~\autoref{fig:cifar10-comparison}, recent methods, including sliding window sampling~\citep{choi2023bws, zheng2024elfs}, message passing~\citep{maharana2023d2}, and stratified sampling (CCS)~\cite{zheng2022coverage}, incorporate mechanisms to exclude a top fraction of the most difficult samples. These designs are motivated by empirical evidence that extreme outliers, whether mislabeled or too ambiguous, can hurt generalization.

While effective on balanced datasets like CIFAR-10/100~\citep{krizhevsky2009learning}, these global cutoff strategies can be brittle in real-world domains with skewed class distributions and uneven difficulty profiles. In network intrusion detection (NID), for instance, malicious traffic is inherently rare and often fragmented across diverse attack types. As shown in~\autoref{tab:cicids2017}, benign traffic accounts for over 80\% of CICIDS2017, while some attack classes (e.g., Heartbleed, Infiltration) contain fewer than 50 examples each.

As shown in Figure~\ref{fig:aum_combined}, AUM scores in CIFAR-10 produce broadly overlapping distributions across classes. In contrast, CICIDS2017 reveals clear class-difficulty separability: attack classes form distinct clusters in difficulty space. The DoS-GoldenEye class, for example, exhibits multiple peaks, likely reflecting temporal or behavioral subpatterns. A global hardest-example threshold in such settings can be harmful; applying a 1\% cutoff rate entirely removes the Web Attack – XSS class. This motivates the need for class-aware sampling strategies that preserve rare but informative examples.

\subsection{Quantifying Class-Wise Difficulty Separability}
\label{ssec:separability}
\begin{table}[t]
\centering
\begin{tabular}{lr}
\toprule
\textbf{Label} & \textbf{\# Samples} \\
\midrule
BENIGN                        & 2,273,097 \\
DoS Hulk                      & 231,073 \\
PortScan                      & 158,930 \\
DDoS                          & 128,027 \\
DoS GoldenEye                 & 10,293 \\
FTP-Patator                   & 7,938 \\
SSH-Patator                   & 5,897 \\
DoS Slowloris                 & 5,796 \\
DoS Slowhttptest              & 5,499 \\
Bot                           & 1,966 \\
Web Attack – Brute Force      & 1,507 \\
Web Attack – XSS              & 652 \\
Infiltration                  & 36 \\
Web Attack – SQL Injection    & 21 \\
Heartbleed                    & 11 \\
\midrule
Total                         & 2,828,743 \\
\bottomrule
\end{tabular}
\caption{Class distribution in the CICIDS2017~\citep{cicids2017} dataset.}
\label{tab:cicids2017}
\end{table}

To quantify the extent to which difficulty scores vary across classes, we introduce the Class Difficulty Separability Coefficient (CDSC). This metric captures how distinguishable the class-conditional distributions of difficulty scores are, based on training dynamics such as AUM or forgetting score. CDSC is defined as a normalized Jensen–Shannon divergence (JSD)~\cite{lin2002divergence} between class distributions, where each class is weighted equally regardless of size. JSD measures the similarity between probability distributions; its normalization ensures CDSC ranges from $[0, 1]$, ensuring fair comparisons across datasets with different numbers of classes. Higher values indicate greater divergence between class difficulty profiles, indicating stronger class-difficulty separability.
\begin{table*}[t]
\centering
\begin{tabular}{p{2.5cm} p{6.6cm} r l p{3cm}}
\toprule
Dataset & Classes / Categories & \# Samples & Domain & Task \\
\midrule
CICIDS2017~\citep{cicids2017} & 1 benign + 14 attacks (e.g., DoS, DDoS) & 2,830,743 & Security & Intrusion Detection \\
CTU‑13~\citep{ctu13} & 1 normal + 13 botnet scenarios & 92,212 & Security & Botnet Detection \\
UNSW‑NB15~\citep{unswnb15} & 1 benign + 9 attacks (e.g., Exploits, Worms) & 5,153 & Security & Intrusion Detection \\
BloodMNIST~\citep{medmnist} & 8 blood cell types (e.g., Eosinophil, Neutrophil) & 17,092 & Medical & Cell Classification \\
DermaMNIST~\citep{medmnist} & 7 skin lesion types (e.g., Melanoma, Nevus) & 10,015 & Medical & Lesion Classification \\
\bottomrule
\end{tabular}
\caption{Overview of datasets used in this study across security and medical domains.}
\label{tab:datasets-overview}
\end{table*}

\begin{algorithm}[t]
\caption{Class-Proportional Sampling}
\label{algo:dist_sampling}
\KwIn{
$\mathcal{S} = \{(x_i, y_i, s_i)\}_{i=1}^n$: dataset with class labels and scores; \\
\hspace{2em} $\alpha$: total coreset sampling rate; \\
\hspace{2em} $m$: minimum number of examples per class; \\
\hspace{2em} $f$: sampling function that takes a data subset, target budget, and optionally a cutoff rate $\beta$
}
\BlankLine
Initialize coreset $\mathcal{S}_c \leftarrow \emptyset$ \;
Total budget $B \leftarrow \lfloor n \times \alpha \rfloor$ \;
$\mathcal{C} \leftarrow$ set of unique classes in $\mathcal{S}$ \;
Let $n_c$ be the number of examples in class $c$ %\tcp*{for all $c \in \mathcal{C}$}
Compute raw budget per class: $B'_c \leftarrow \max(\lfloor B \cdot n_c / n \rfloor, m)$ for all $c$ \;
\If{$\sum_{c \in \mathcal{C}} B'_c > B$}{
    Sort classes by $n_c$ in descending order \;
    \While{$\sum_{c \in \mathcal{C}} B'_c > B$}{
        \ForEach{class $c$ in sorted $\mathcal{C}$}{
            \If{$B'_c > m$}{
                $B'_c \leftarrow B'_c - 1$ \;
                \If{$\sum B'_c \leq B$}{\textbf{break}}
            }
        }
    }
}
\BlankLine
\ForEach{class $c \in \mathcal{C}$}{
    $\mathcal{S}^c \leftarrow \{(x_i, y_i, s_i) \in \mathcal{S} \mid y_i = c\}$ \tcp*{Class-specific subset}
    
    $S_c \leftarrow f(\mathcal{S}^c, B'_c)$ \;

    $\mathcal{S}_c \leftarrow \mathcal{S}_c \cup S_c$ \;
}
\Return $\mathcal{S}_c$
\end{algorithm}

\noindent\textbf{Class Difficulty Separability Coefficient (CDSC).}
Let $\mathcal{X} = \{x_1, \ldots, x_N\}$ be a dataset with class labels $y_i \in \{1, \ldots, C\}$,
and let $s(x_i) \in \mathbb{R}$ denote a scalar difficulty score for example $x_i$ derived from training dynamics.
For each class $c \in \{1, \ldots, C\}$, let $\mathcal{X}_c = \{x_i \in \mathcal{X} \mid y_i = c\}$ denote the set of examples from class $c$,
and let $P_c$ be the normalized probability distribution over the scores $\{s(x_i) \mid x_i \in \mathcal{X}_c\}$, estimated via kernel density estimation (KDE)~\cite{davis2011remarks}. Define the mixture distribution as $M = \frac{1}{C} \sum_{c=1}^C P_c$, assigning equal weight to each class.
We define the \emph{Class Difficulty Separability Coefficient}, denoted by $\delta_{\text{CD}}$, as:

\begin{equation}
\delta_{\text{CD}} := \frac{1}{\log_2 C} \left( H\left(M\right) - \frac{1}{C} \sum_{c=1}^C H(P_c) \right),
\end{equation}

where $H(P) = -\sum_j P_j \log P_j$ is the Shannon entropy of distribution $P$.
The coefficient $\delta_{\text{CD}}$ takes the value $0$ when all class-conditional difficulty distributions are identical, 
and approaches $1$ as their supports become mutually disjoint.

We use $\delta_{\text{CD}}$ as a quantitative indicator of how separable the class-wise difficulty distributions are within a dataset. In Section~\ref{sssec:cdsc}, we show that datasets with higher $\delta_{\text{CD}}$ tend to benefit more from class-aware coreset selection, achieving larger performance gains in accuracy, precision, and recall. This supports our hypothesis that accounting for class difficulty structure is especially important when such separability is high.

\subsection{Class-Proportional Coreset Selection}

To address the limitations of standard sampling strategies under class imbalance, we introduce a class-balanced variant applicable to a range of difficulty-based coreset selection methods. This variant serves as a simple but effective patch to commonly used techniques, including hardest-example-only sampling~\cite{pleiss2020identifying, toneva2018empirical}, hard-cutoff with stratified sampling (Coverage-Centric Selection, CCS)~\cite{zheng2022coverage}, and sliding window sampling~\cite{choi2023bws, zheng2024elfs}. As shown in Algorithm~\ref{algo:dist_sampling}, instead of applying sampling over the entire data difficulty distribution, we perform sampling within each class independently. This ensures that the sampling budget is allocated more evenly across classes, preventing rare classes from being overlooked. This modification improves robustness in highly imbalanced settings without significantly altering the underlying selection strategy.

\section{Experiment Setup}

\subsection{Datasets}

\textbf{Security Datasets.} 
As shown in~\autoref{tab:datasets-overview}, we use three widely adopted network intrusion detection datasets and two medical imaging datasets. 
For security, we include: 
1) CICIDS2017~\cite{cicids2017}, which includes a range of realistic attack scenarios such as brute force, DDoS, infiltration, and web-based attacks, captured alongside benign traffic in a controlled enterprise-like environment;
2) UNSW-NB15~\cite{unswnb15}, which provides rich feature sets extracted from traffic generated in a testbed simulating both normal operations and diverse modern attacks;
and 
3) CTU-13~\cite{ctu13}, which captures real botnet infections in NetFlow format and presents significant class imbalance.
Together, these datasets offer a diverse and representative benchmark for evaluating data-efficient NID methods.

\textbf{Medical Datasets.} 
In the medical domain, we include two subsets from the MedMNIST collection~\cite{medmnist}: 
4) BloodMNIST, which contains 8 classes of blood cell types (e.g., eosinophils, neutrophils), and 
5) DermaMNIST, a skin lesion classification dataset with 7 categories including melanoma and nevus. 
These datasets exhibit varying levels of class imbalance and intra-class diversity, offering complementary perspectives for evaluating class-difficulty separability in non-security domains. More details about the datasets used in this study can be found in~\autoref{app:datasets}.

\begin{figure*}[ht!]
\centering

% Row 1: CTU13
\begin{subfigure}[b]{0.3\textwidth}
  \centering
  \includegraphics[width=\textwidth]{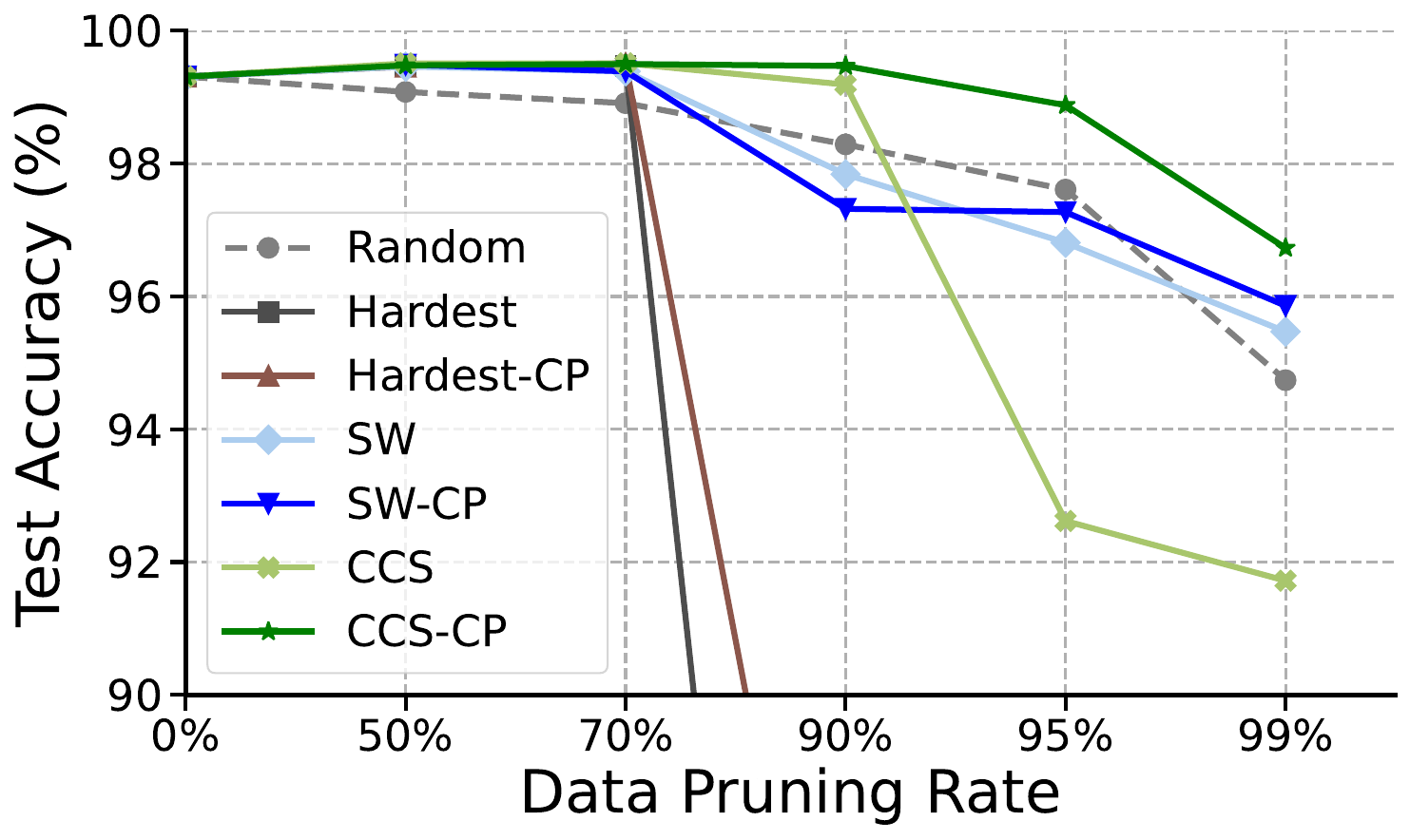}
  \caption{CTU13 Accuracy.}
  \label{fig:ctu13-acc}
\end{subfigure}
\hspace{0.5em}
\begin{subfigure}[b]{0.3\textwidth}
  \centering
  \includegraphics[width=\textwidth]{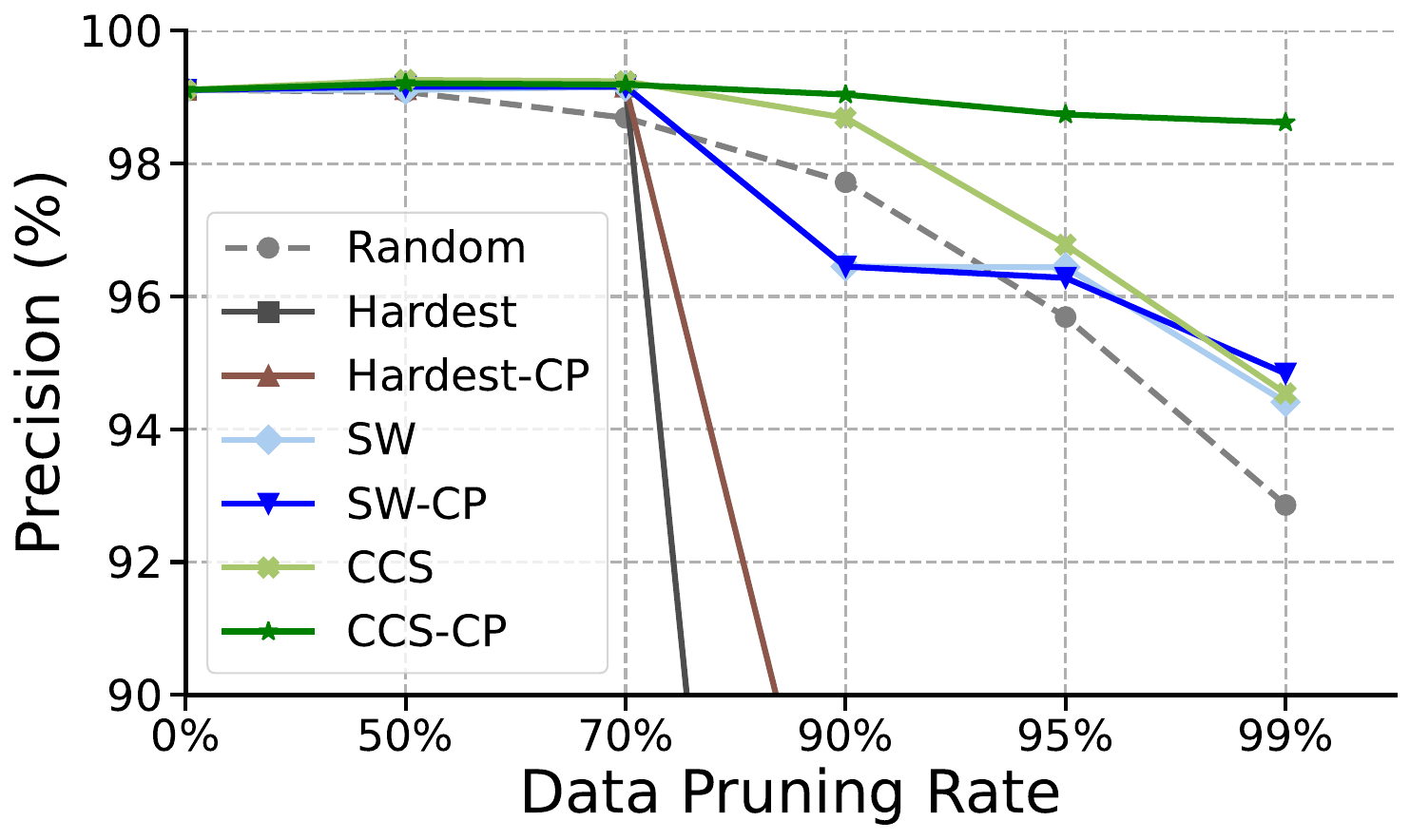}
  \caption{CTU13 Precision.}
  \label{fig:ctu13-prec}
\end{subfigure}
\hspace{0.5em}
\begin{subfigure}[b]{0.3\textwidth}
  \centering
  \includegraphics[width=\textwidth]{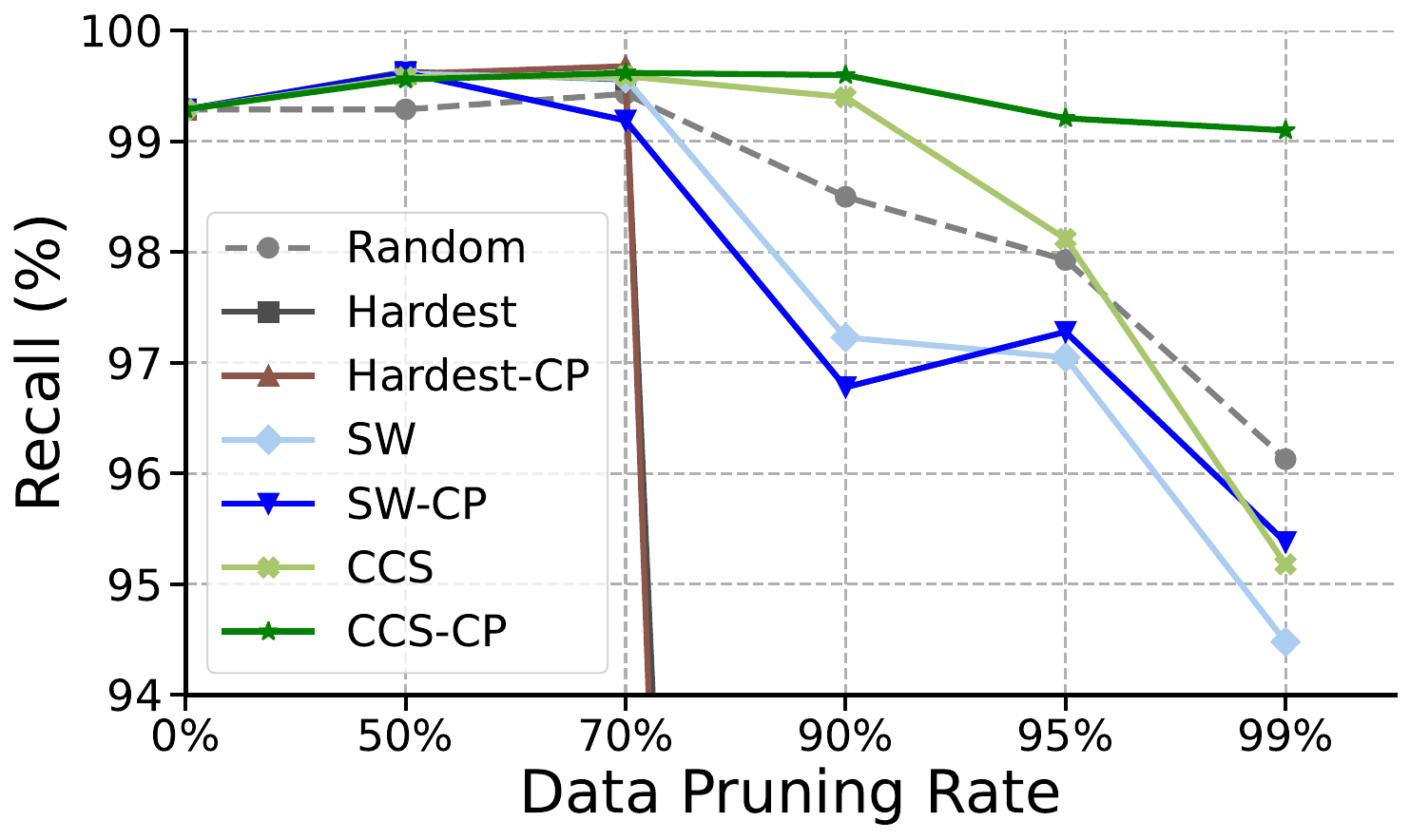}
  \caption{CTU13 Recall.}
  \label{fig:ctu13-recall}
\end{subfigure}

\vspace{1.5em}

% Row 2: UNSW-NB15
\begin{subfigure}[b]{0.3\textwidth}
  \centering
  \includegraphics[width=\textwidth]{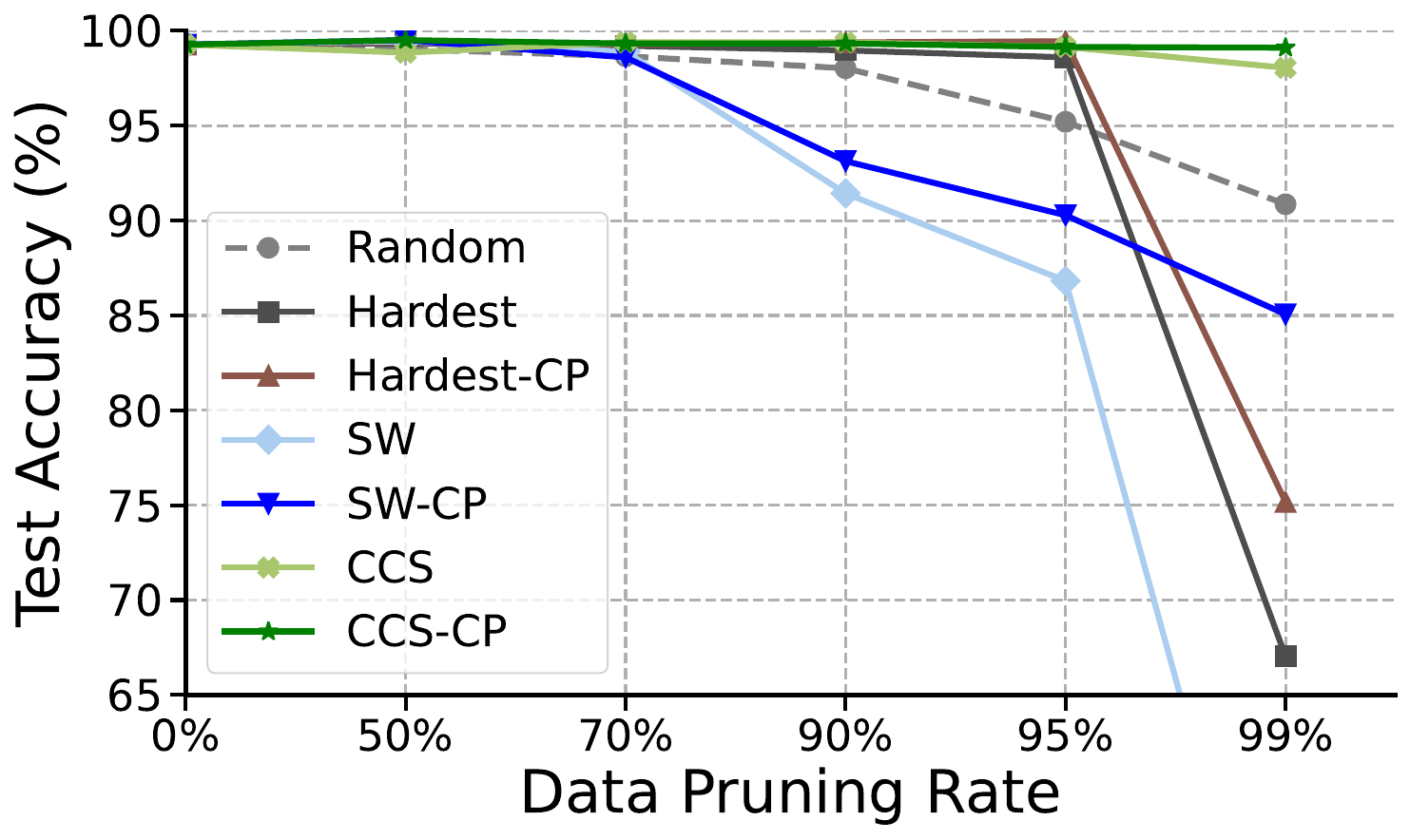}
  \caption{UNSW-NB15 Accuracy.}
  \label{fig:unsw-acc}
\end{subfigure}
\hspace{0.5em}
\begin{subfigure}[b]{0.3\textwidth}
  \centering
  \includegraphics[width=\textwidth]{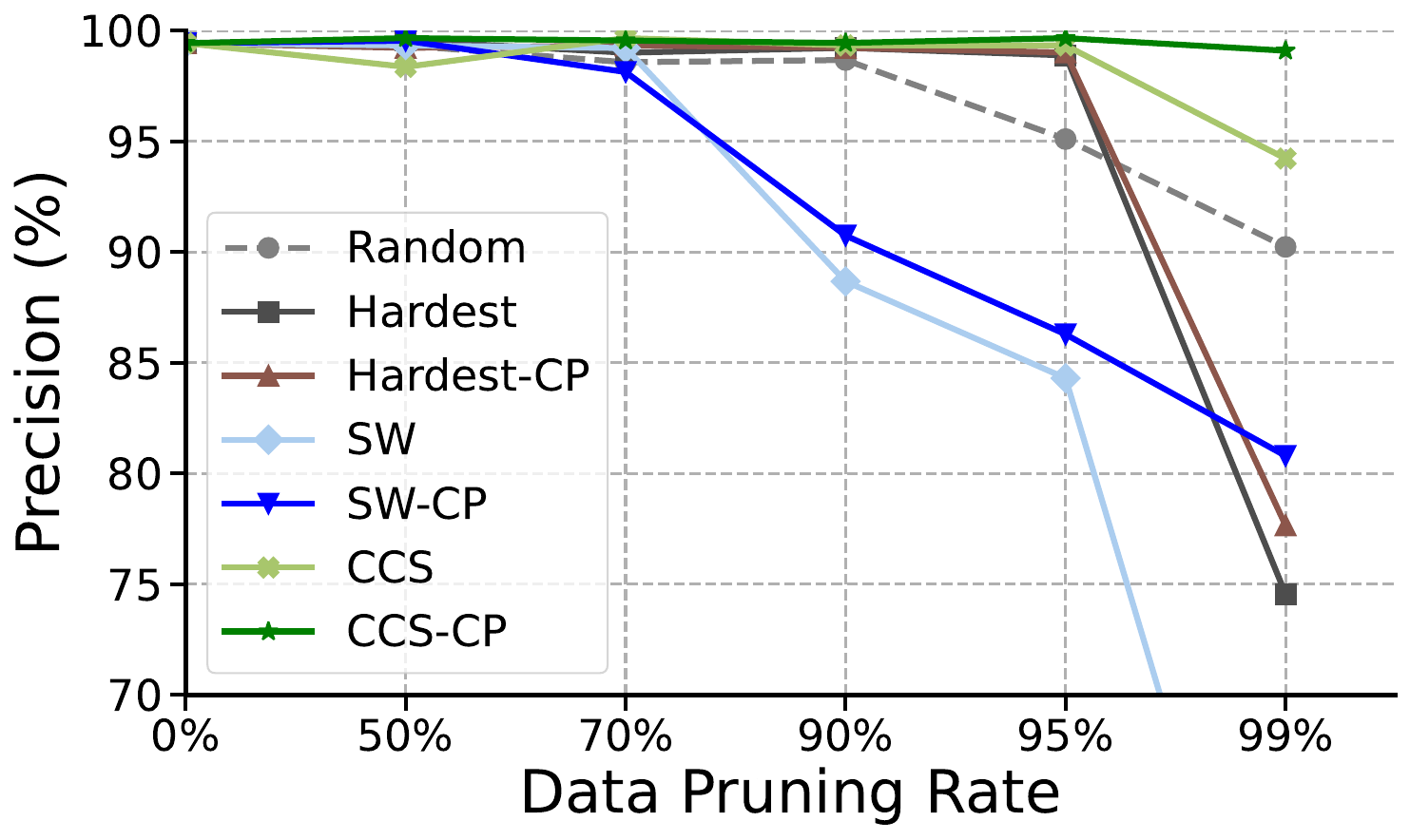}
  \caption{UNSW-NB15 Precision.}
  \label{fig:unsw-prec}
\end{subfigure}
\hspace{0.5em}
\begin{subfigure}[b]{0.3\textwidth}
  \centering
  \includegraphics[width=\textwidth]{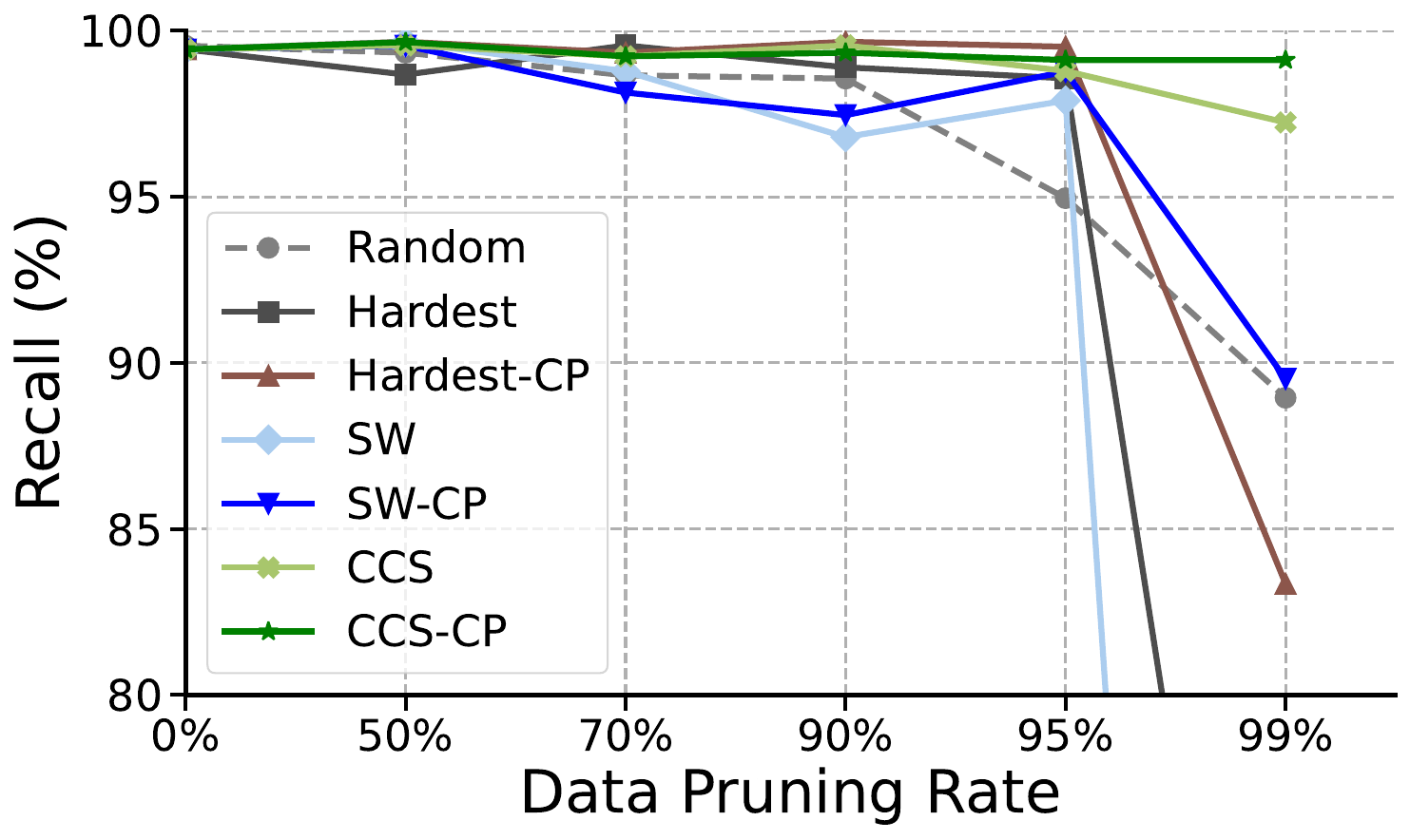}
  \caption{UNSW-NB15 Recall.}
  \label{fig:unsw-recall}
\end{subfigure}

\vspace{1.5em}

% Row 3: CICIDS2017
\begin{subfigure}[b]{0.3\textwidth}
  \centering
  \includegraphics[width=\textwidth]{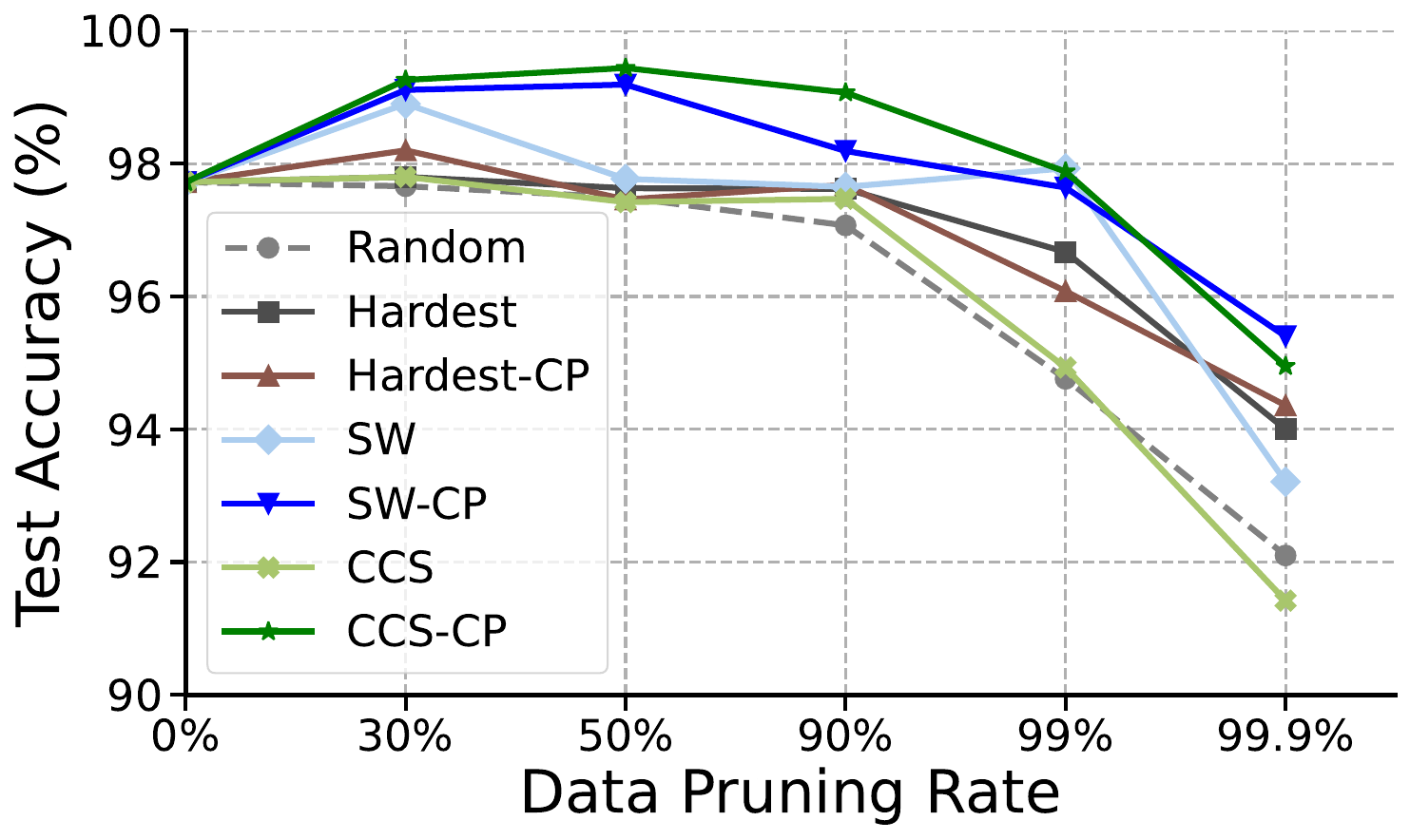}
  \caption{CICIDS2017 Accuracy.}
  \label{fig:cicids-acc}
\end{subfigure}
\hspace{0.5em}
\begin{subfigure}[b]{0.3\textwidth}
  \centering
  \includegraphics[width=\textwidth]{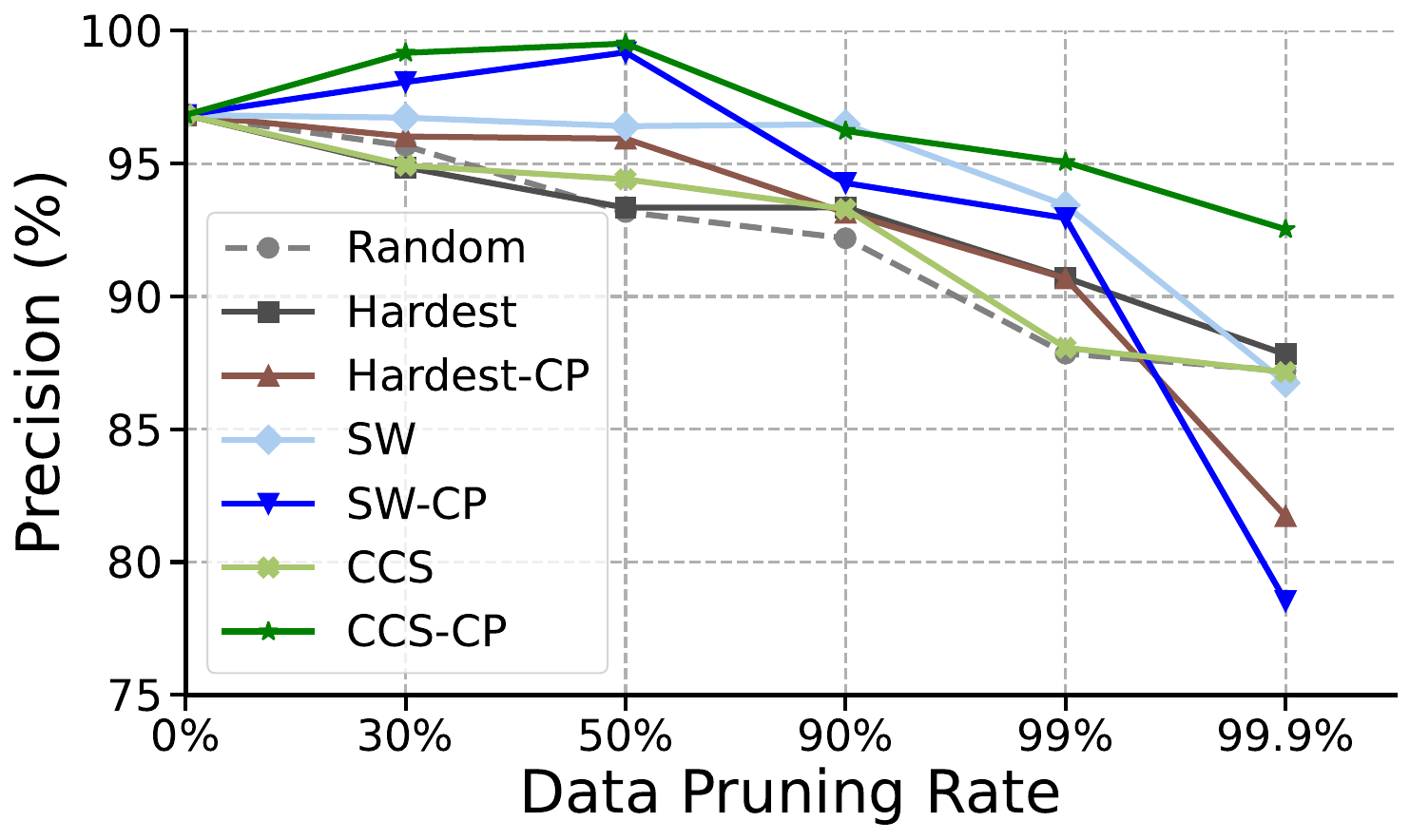}
  \caption{CICIDS2017 Precision.}
  \label{fig:cicids-prec}
\end{subfigure}
\hspace{0.5em}
\begin{subfigure}[b]{0.3\textwidth}
  \centering
  \includegraphics[width=\textwidth]{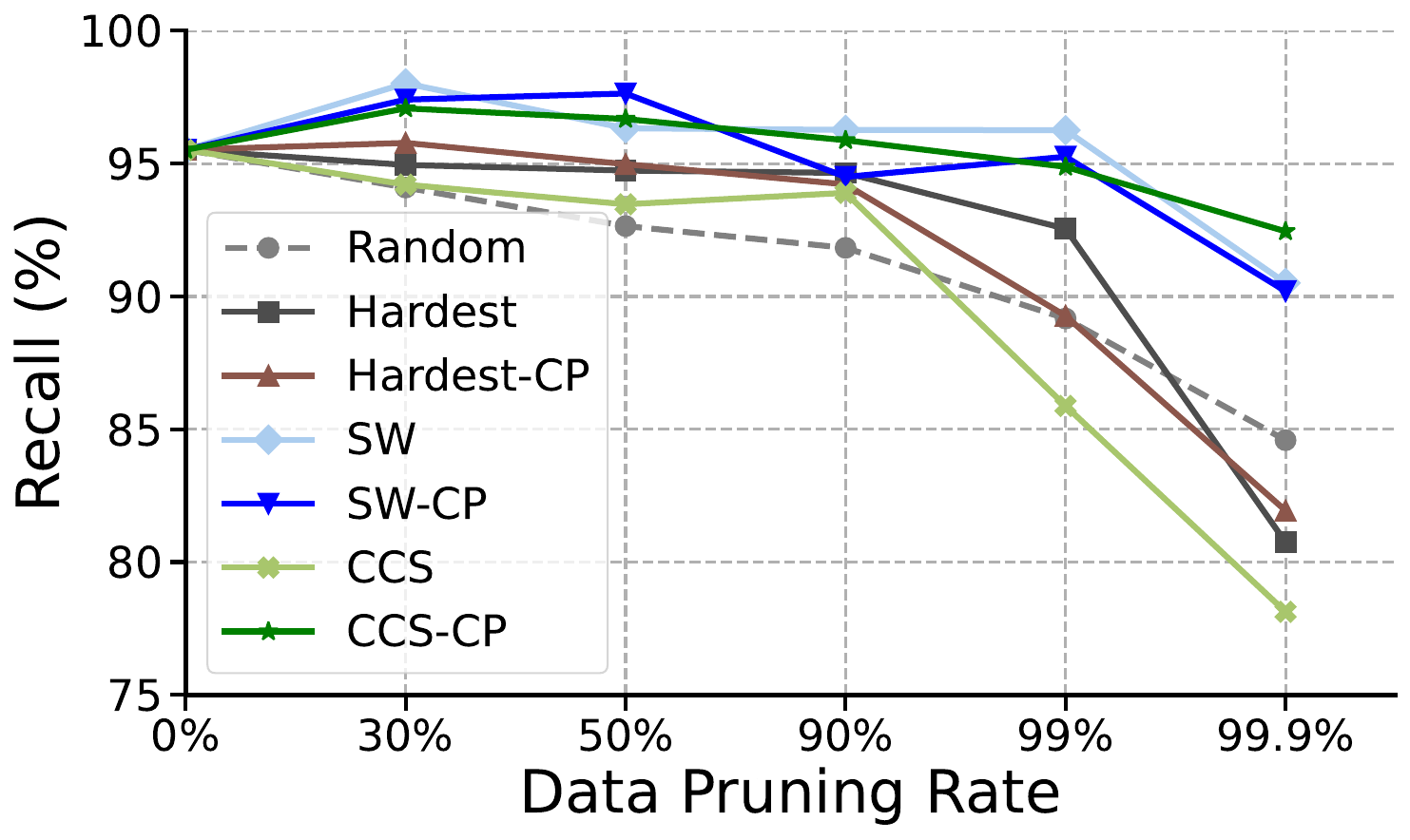}
  \caption{CICIDS2017 Recall.}
  \label{fig:cicids-recall}
\end{subfigure}

\vspace{1.5em}

% Row 4: DermaMNIST and BloodMNIST Accuracy
\begin{subfigure}[b]{0.3\textwidth}
  \centering
  \includegraphics[width=\textwidth]{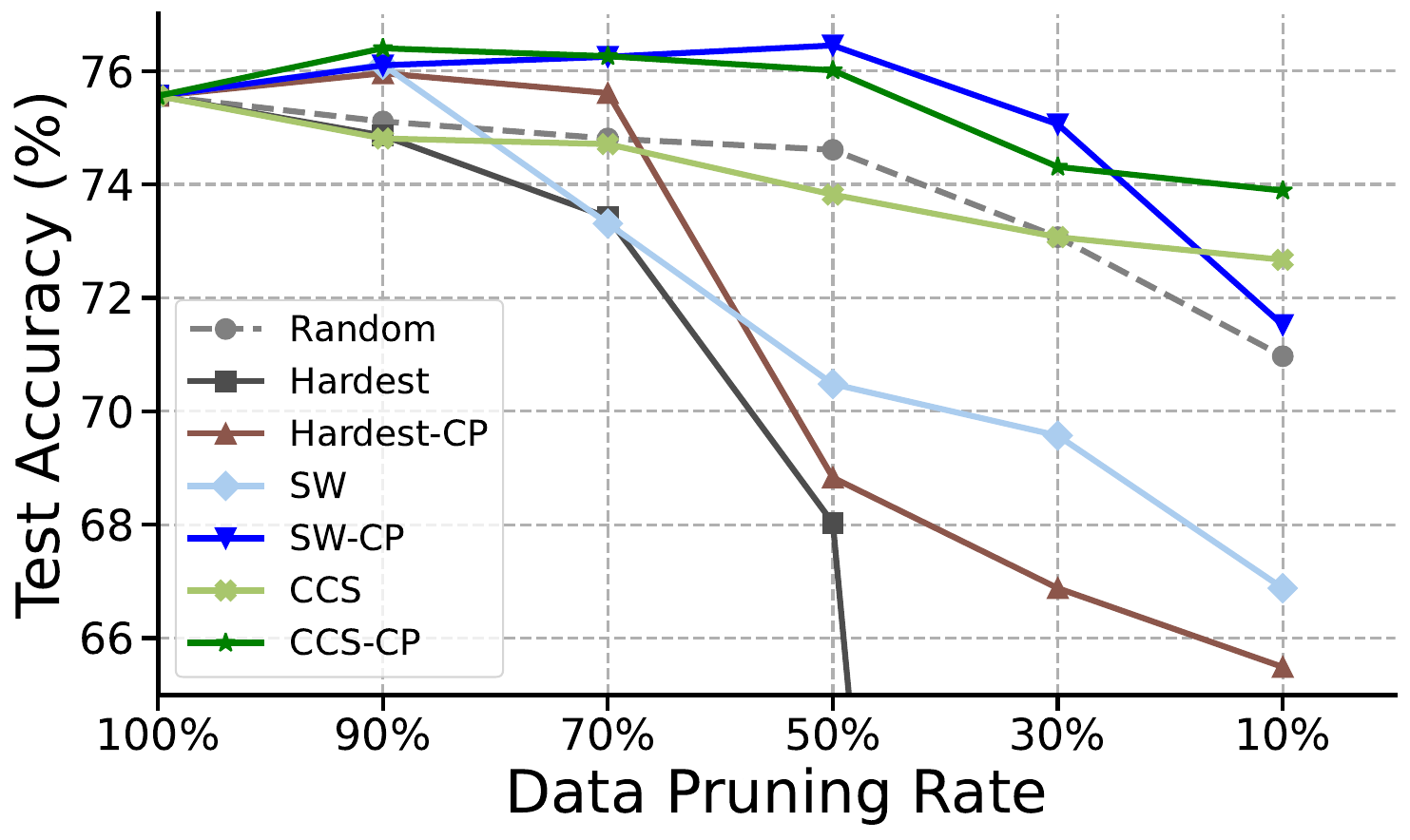}
  \caption{DermaMNIST Accuracy.}
  \label{fig:derma-acc}
\end{subfigure}
\hspace{2em}
\begin{subfigure}[b]{0.3\textwidth}
  \centering
  \includegraphics[width=\textwidth]{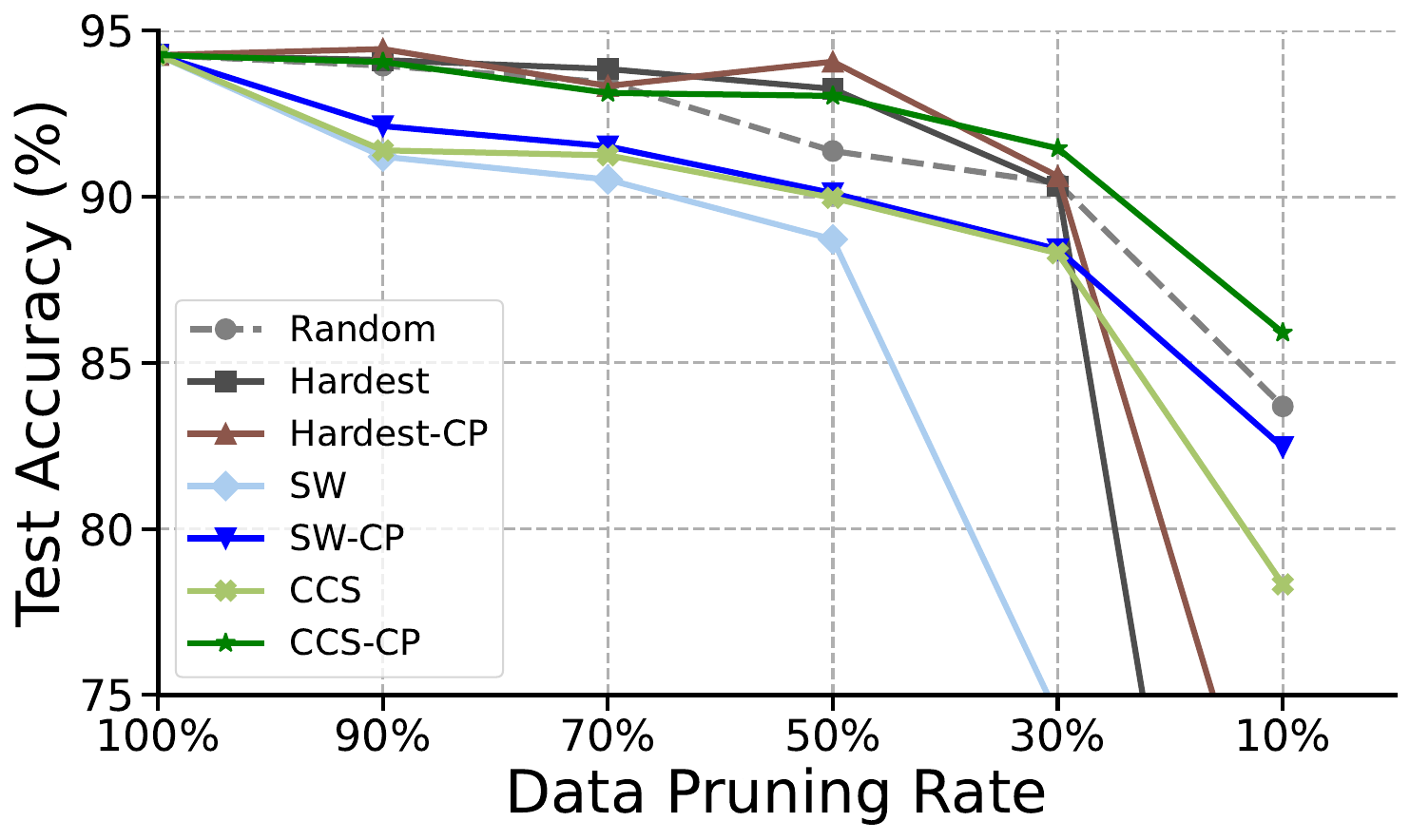}
  \caption{BloodMNIST Accuracy.}
  \label{fig:blood-acc}
\end{subfigure}

\vspace{-0.5em}
\caption{Performance comparison across random pruning and other data selection baselines on benchmark datasets: CTU13~\cite{ctu13}, UNSW-NB15~\cite{unswnb15}, CICIDS2017~\cite{cicids2017}, DermaMNIST~\cite{medmnist}, and BloodMNIST~\cite{medmnist}.}
\label{fig:performance-comparison}
\end{figure*}

%\vspace{1em}
\textbf{Dataset Preprocessing.} For UNSW-NB15 and CTU-13, we follow prior work in applying flow-based aggregation to better capture high-level traffic patterns. The statistics reported in~\autoref{tab:datasets-overview} reflect the datasets after preprocessing and aggregation.
Specifically, for UNSW-NB15, we follow Gwon et al.~\cite{gwon1911network}, segmenting flows into sequences of 50 using a sliding window and assigning binary labels via majority voting for sequence-level classification. For CTU-13, we use the preprocessed dataset from Hussain et al.~\cite{hussain2020towards}, which applies CICFlowMeter~\cite{draper2016characterization,lashkari2017cicflowmeter} to extract flow-level features. 

\subsection{Baselines and Evaluation Metrics} 
We compare our class-proportional variants against several widely used coreset selection strategies:
1) \textbf{Random} samples data uniformly at random and serves as a naive, class-agnostic baseline;
2) \textbf{Hardest}~\cite{pleiss2020identifying, toneva2018empirical} selects the most difficult examples according to a training-dynamics–based difficulty score (e.g., AUM or forgetting score);
3) \textbf{Sliding Window (SW)}~\cite{choi2023bws, zheng2024elfs} selects examples from an interval $[\beta, \beta + \alpha]$ of the sorted difficulty scores, where $\alpha$ is the target pruning rate and $\beta$ is a tunable starting offset;
4) \textbf{Coverage-Centric Sampling (CCS)}~\cite{zheng2022coverage} discards the top $\beta\%$ hardest examples and then applies stratified sampling over the remaining data to select an $\alpha$-fraction coreset.

For each of these methods, we implement a \textit{class-proportional (CP)} variant—\textbf{Hardest-CP}, \textbf{SW-CP}, and \textbf{CCS-CP}—by applying the same selection procedure independently within each class and allocating the sampling budget proportionally.

\textbf{Evaluation Metrics.} 
For security datasets, we report test accuracy, as well as precision and recall computed over the attack classes. This is important because these datasets are typically imbalanced, and accuracy alone may mask poor detection performance on rare but critical attack types. 
For medical datasets, where each class corresponds to a clinically meaningful condition, we report overall test accuracy. In these settings, all classes are of interest, and class imbalance is generally less extreme.

\begin{table*}[ht]
\centering
\begin{tabular}{lccccc}
\toprule
\textbf{Metric} & UNSW-NB15 & CICIDS2017 & CTU13 & DermaMNIST & BloodMNIST \\
\midrule
$\delta_{\text{CD}}$        & 0.067  & 0.114   & 0.489  & 0.082  & 0.280  \\
\midrule
$\Delta$ Accuracy (\%)      & +1.06  & +3.51   & +5.01  & +1.21  & +7.57  \\
$\Delta$ Precision (\%)     & +4.87  & +5.37   & +4.08  & --     & --     \\
$\Delta$ Recall (\%)        & +1.98  & +14.32  & +3.91  & --     & --     \\
\bottomrule
\end{tabular}
\caption{
Class Difficulty Separability Coefficient ($\delta_{\text{CD}}$) quantifies the divergence between class-specific AUM score distributions, capturing how separable class difficulty profiles are. The table also reports absolute performance gains (in percentage points) of class-proportional CCS (CCS-CP) over standard CCS at the highest pruning rate in~\autoref{fig:performance-comparison}. 
}
\label{tab:performance_boost}
\end{table*}

\subsection{Implementation}
We use task-specific model architectures for each dataset. For UNSW-NB15, we adopt an LSTM-based architecture, following prior work on this dataset~\cite{gwon1911network}. For CTU-13, we implement the hybrid ResDN+ViT model (ResDNViT) proposed by Wasswa et al.~\cite{wasswa2025resdnvit}. For CICIDS2017, DermaMNIST, and BloodMNIST, we use a lightweight CNN architecture. We find that larger models, such as ResNet18~\cite{he2016deep}, lead to rapid convergence, making it difficult to collect meaningful training dynamics. Therefore, we favor simpler models that expose more informative learning signals across epochs. Following prior work~\cite{zheng2024elfs}, we use the Area Under the Margin (AUM) score to quantify data difficulty. More details can be found in~\autoref{app:implementation}.

\section{Evaluation \& Discussion}

\subsection{Main Evaluation}

To assess the impact of class-aware sampling, we compare class-proportional coreset selection methods with class-agnostic baselines across five benchmark datasets in~\autoref{fig:performance-comparison}. Our proposed class-proportional variants—Hardest-CP, SW-CP, and CCS-CP—consistently outperform their class-agnostic counterparts, especially at aggressive pruning rates ($\geq
90\%$). This performance gap is most pronounced on security datasets like CTU13, UNSW-NB15, and CICIDS2017, where CCS-CP achieves superior precision and recall while maintaining high accuracy even after removing 90–99.9\% of the training data. On DermaMNIST and BloodMNIST, CCS-CP also shows resilience at lower data volumes, with accuracy remaining stable while class-agnostic methods degrade sharply. We present and analyze a few representative
sub-populations selected by CCS-CP in~\autoref{fig:visualization}. Across all datasets, class-proportional variants yield flatter performance curves under pruning, indicating improved robustness in retaining informative and balanced samples.

On security datasets with high class-difficulty separability (see Section~\ref{sssec:cdsc}), CCS-CP consistently delivers the strongest performance across accuracy, precision, and recall---especially under aggressive pruning. On CTU-13, CCS-CP matches the full-data accuracy at 90\% pruning and even improves recall ($+0.53\%$) while keeping the precision drop minimal ($-0.07\%$). At an extreme $99\%$ pruning rate, CCS-CP maintains remarkable stability, with only a $2.58\%$ drop in accuracy, $0.49\%$ in precision, and $0.19\%$ in recall. In contrast, the class-agnostic CCS baseline---the next best method in terms of overall accuracy, precision, and recall---suffers much steeper declines: $7.59\%$ in accuracy, $4.57\%$ in precision, and $4.11\%$ in recall. 
Similar patterns hold across UNSW-NB15, CICIDS2017, DermaMNIST, and BloodMNIST, where CCS-CP maintains stronger accuracy, precision, and recall even under high pruning rates. This underscores the strength of class-proportional selection in preserving high-quality, representative examples from all classes. In contrast, class-agnostic methods may inadvertently discard critical minority classes, leading to degraded performance, particularly harmful for imbalanced datasets common in medical image classification and security domains such as network intrusion detection.

\textbf{Coreset Selection to Improve Generalization.}
Toneva et al.~\cite{toneva2018empirical} showed that removing up to 30\% of the CIFAR-10 training data, specifically the examples that are never forgotten during training, can maintain or even slightly improve test accuracy. We observe a similar effect in our experiments. On both network intrusion detection datasets such as CICIDS2017 and medical image datasets such as DermaMNIST, test accuracy, precision, and recall sometimes improve when training on carefully selected subsets instead of the full dataset. This improvement is possible because many training examples are either too easy (and thus redundant) or too hard (potentially acting as outliers that lead to overfitting). Removing these examples focuses training on more informative samples, leading to better generalization. These results collectively underscore the importance of accounting for class-difficulty separability when designing data-efficient learning pipelines.

\subsection{Ablation Study and Analysis}
\subsubsection{Class-Wise Difficulty Separability}
\label{sssec:cdsc}

\autoref{tab:performance_boost} presents the Class Difficulty Separability Coefficient ($\delta_{\text{CD}}$) for each dataset, along with the performance improvements achieved by our class-proportional CCS over the standard CCS baseline at the highest evaluated pruning rate. As a baseline, CIFAR-10 exhibits a $\delta_{\text{CD}}$ of $0.043$, indicating low class-difficulty separability. In contrast, datasets such as CTU-13 ($\delta_{\text{CD}} = 0.489$) and BloodMNIST ($\delta_{\text{CD}} = 0.280$) show significantly higher separability, and correspondingly larger improvements in accuracy ($5.01\%$ and $7.57\%$, respectively). Meanwhile, datasets with lower separability, such as UNSW-NB15 and DermaMNIST, yield smaller gains. These results support our hypothesis that class-aware sampling is especially beneficial when difficulty varies substantially across classes. Notably, on CICIDS2017 CCS-CP achieves a $14.32\%$ gain in recall when pruning $99.9\%$ of the data, illustrating the advantage of preserving rare but informative examples that are often pruned by class-agnostic strategies.

\subsubsection{Data Difficulty Metrics}

\begin{table}[h]
\centering
\begin{tabular}{lccccc}
\toprule
Metric & 10\% & 30\% & 50\% & 70\% & 90\% \\
\midrule
AUM        & 99.48 & \textbf{99.50} & 99.08 & \textbf{98.88} & \textbf{96.73} \\
Forgetting & 99.41 & \textbf{99.50} & \textbf{99.43} & 97.70 & 95.93 \\
EL2N       & \textbf{99.67} & 99.10 & 99.21 & 98.71 & 95.66 \\
\bottomrule
\end{tabular}
\caption{Test accuracy (\%) on UNSW-NB15 using CCS-CP at different pruning rates with three difficulty metrics.}
\label{tab:metric_comparison}
\end{table}

\autoref{tab:metric_comparison} shows that CCS-CP, when used with different training dynamics–based difficulty metrics, achieves very similar coreset selection performance across all pruning rates. This suggests that the specific choice of scoring metric, such as AUM~\cite{pleiss2020identifying}, Forgetting~\cite{toneva2018empirical}, and EL2N~\cite{paul2021deep}, has only a marginal impact on overall effectiveness. All three metrics enable CCS-CP to maintain high test accuracy, particularly at moderate pruning levels, indicating that the method is robust to the form of the difficulty metric used for coreset selection.

\subsubsection{Coreset Qualitative Analysis}

\begin{figure}[ht]
\centering
\includegraphics[width=0.9\linewidth]{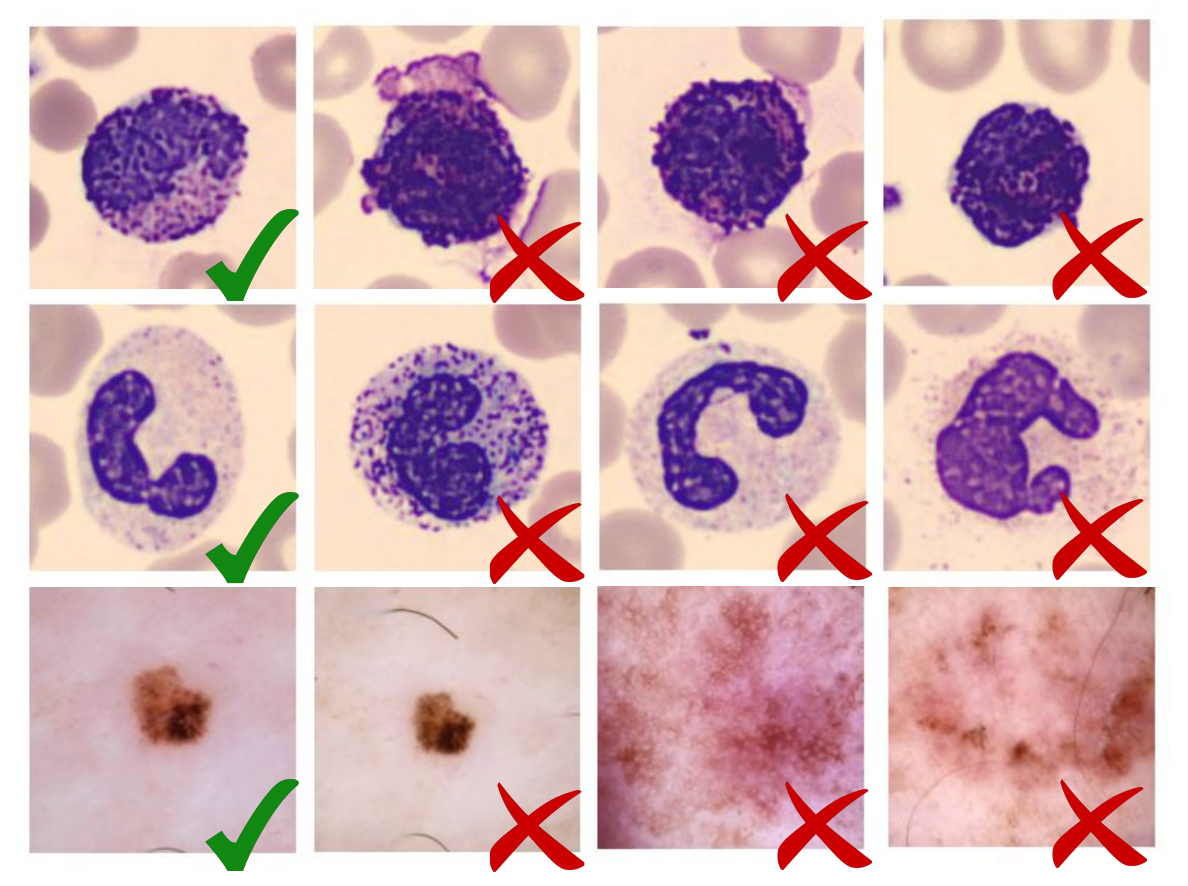}
\caption{
Example of coresets selected by CCS-CP at a 10\% pruning rate.
Each row corresponds to a different class: the first row shows \textbf{Basophil} from BloodMNIST, the second row shows \textbf{Neutrophil} from BloodMNIST, and the third row shows \textbf{Melanocytic nevi} from DermaMNIST.
Images are grouped by class to highlight intra-class selection behavior.
}
\label{fig:visualization}
\end{figure}

\autoref{fig:visualization} illustrates CCS-CP’s intra-class selection behavior by visualizing retained (\checkmark) and pruned ($\times$) samples across three classes. In each row, CCS-CP tends to preserve representative and clean images while discarding visually redundant or ambiguous cases. For example, in the Basophil and Neutrophil rows (BloodMNIST), retained cells exhibit clear morphological features and consistent staining, whereas discarded ones are blurry, occluded, or similar to already selected samples. In the Melanocytic nevi row (DermaMNIST), the selected image shows a distinct and well-defined lesion, while the pruned images are either low-contrast or similar to the selected one. This suggests CCS-CP effectively identifies both prototypical and diverse examples while pruning out low-quality or duplicative data.
\section{Conclusion}

We investigate the limitations of class-agnostic coreset selection methods in real-world settings where class-wise data difficulty varies significantly. We introduce the concept of \textit{class-difficulty separability}, quantify it using the Class Difficulty Separability Coefficient (CDSC), and show that it correlates with the failure modes of standard coreset methods. To address this, we propose class-proportional variants of existing sampling strategies, which allocate the sampling budget proportionally across classes and operate on difficulty scores within each class.

Extensive experiments across five datasets---including three network intrusion detection benchmarks and two medical imaging tasks---demonstrate that our class-proportional methods significantly outperform their class-agnostic counterparts, particularly under high pruning rates. In some cases, they even improve generalization over full-data training, highlighting their utility in noisy or imbalanced domains.

Our findings suggest that accounting for class-difficulty structure is critical for effective and trustworthy data pruning in practice. We hope this work inspires further exploration of class-aware data selection and its theoretical foundations.

{
    \small
    \bibliographystyle{ieeenat_fullname}
    \bibliography{main}
}
\newpage

\appendix
\section*{Appendix}
\label{sec:appendix}

\section{Dataset Details}
\label{app:datasets}

This appendix provides detailed information about the class labels and structure of each dataset used in our experiments.

\subsection{Network Intrusion Detection Datasets}

\textbf{CICIDS2017}~\cite{cicids2017} includes a total of 15 classes:
\begin{itemize}
    \item \textbf{Benign}.
    \item \textbf{Attacks:} FTP-Patator, SSH-Patator, DoS-GoldenEye, DoS-Hulk, DoS-SlowHTTPTest, DoS-Slowloris, Heartbleed, Web Attack–Brute Force, Web Attack–XSS, Web Attack–SQL Injection, Infiltration, Bot, PortScan, DDoS.
\end{itemize}
\vspace{1em}
\noindent\textbf{UNSW-NB15}~\cite{unswnb15} contains 10 classes:
\begin{itemize}
    \item \textbf{Benign}.
    \item \textbf{Attacks:} Fuzzers, Analysis, Backdoors, DoS, Exploits, Generic, Reconnaissance, Shellcode, Worms.
\end{itemize}

\vspace{1em}
\noindent\textbf{CTU-13}~\cite{ctu13} consists of 13 distinct scenarios capturing botnet activity. Each scenario includes three traffic classes:
\begin{itemize}
    \item \textbf{Botnet} (malicious).
    \item \textbf{Normal}.
    \item \textbf{Background} (unlabeled or unknown).
\end{itemize}
In our setup, we treat each botnet-infected flow as a positive instance and group the Normal and Background flows as negative.

\subsection{Medical Imaging Datasets}

\textbf{BloodMNIST} (from MedMNIST v2~\cite{medmnist}) includes 8 blood cell categories:
\begin{itemize}
    \item Neutrophil segmented.
    \item Eosinophil.
    \item Basophil.
    \item Lymphocyte.
    \item Monocyte.
    \item Immature granulocyte.
    \item Erythroblast.
    \item Platelet.
\end{itemize}

\vspace{1em}
\noindent\textbf{DermaMNIST} contains 7 skin lesion types:
\begin{itemize}
    \item Actinic keratoses.
    \item Basal cell carcinoma.
    \item Benign keratosis-like lesions.
    \item Dermatofibroma.
    \item Melanocytic nevi.
    \item Vascular lesions.
    \item Melanoma.
\end{itemize}

These class-level details help contextualize the class-difficulty separability observed in our experiments.

\section{Implementation}
\label{app:implementation}

All models are trained using the Adam optimizer~\cite{loshchilov2017decoupled} with early stopping based on validation accuracy. We scale the maximum number of training epochs inversely with the coreset sampling fraction to ensure fair comparisons across pruning levels. Learning rates are set to $2\text{e-}4$ for CTU-13~\cite{ctu13} and $1\text{e-}3$ for the rest. Batch sizes are 64 for UNSW-NB15~\cite{unswnb15} and CTU-13~\cite{ctu13}, 128 for DermaMNIST and BloodMNIST~\cite{medmnist}, and 512 for CICIDS2017~\cite{cicids2017}.

All experiments employ early stopping with a patience of 30–50 epochs. The best-performing model on the validation set is saved for evaluation. We evaluate all models under multiple coreset sampling strategies using precomputed training dynamics (e.g., AUM) for difficulty-based selection. 

\subsection{Hierarchical Beta Search}

To identify the optimal $\beta$ parameter for coreset selection methods that rely on this hyperparameter, namely CCS, CCS-CP, Window, and Window-CP, we adopt a hierarchical search strategy that is both efficient and effective. This approach is motivated by an empirical observation, consistent with prior work~\cite{zheng2022coverage}, that the relationship between $\beta$ and the test accuracy is approximately convex. This structure allows us to search efficiently without risking poor local optima.

The search is conducted in three phases, each with a finer step size: $\delta = 0.1$, $0.05$, and $0.01$. These granularities are chosen based on the smallest coreset ratio used in most datasets, which is $0.01$. Each phase begins at the $\beta$ found in the previous phase ($\beta = 0$ for phase $0$) and moves forward by incrementally adding $\delta$. The algorithm continues as long as validation accuracy improves. Once a decline is observed, it backtracks by $1.5 \times \delta$ and proceeds to the next phase. To reduce redundant computation, we cache performance results for previously evaluated $\beta$ values. The search terminates after completing all three phases and returns the $\beta$ value that yielded the best validation accuracy.

This hierarchical scheme typically requires only 10–20 experiments to converge and achieves robust results across different datasets and sampling strategies.

\end{document}